\def\arxivpreprint{}

\documentclass[letterpaper]{article} 
\usepackage[submission]{aaai2026}  
\usepackage{times}  
\usepackage{helvet}  
\usepackage{courier}  
\usepackage[hyphens]{url}  
\usepackage{graphicx} 
\usepackage{natbib}  
\usepackage{caption} 
\usepackage{algorithm}
\usepackage{algorithmic}

\usepackage{newfloat}
\usepackage{listings}
\DeclareCaptionStyle{ruled}{labelfont=normalfont,labelsep=colon,strut=off} 
\floatstyle{ruled}
\newfloat{listing}{tb}{lst}{}
\floatname{listing}{Listing}
\usepackage{subcaption}
\usepackage{makecell}
\usepackage{multirow}
\usepackage{multicol}
\usepackage{amsmath}
\usepackage{amsfonts}

\title{HT-Transformer: Event Sequences Classification by Accumulating Prefix Information with History Tokens}
\author{
    Ivan Karpukhin,
    Andrey Savchenko \\ Sber AI Lab, Russia
}
\affiliations{
    \textsuperscript{\rm 1}Sber AI Lab\\

    Moscow, Russia\\
    i.a.karpukhin@sberbank.ru
}

\usepackage{bibentry}

\begin{document}

\maketitle

\begin{abstract}
Deep learning has achieved remarkable success in modeling sequential data, including event sequences, temporal point processes, and irregular time series. Recently, transformers have largely replaced recurrent networks in these tasks. However, transformers often underperform RNNs in classification tasks where the objective is to predict future targets. The reason behind this performance gap remains largely unexplored. In this paper, we identify a key limitation of transformers: the absence of a single state vector that provides a compact and effective representation of the entire sequence. Additionally, we show that contrastive pretraining of embedding vectors fails to capture local context, which is crucial for accurate prediction. To address these challenges, we introduce history tokens, a novel concept that facilitates the accumulation of historical information during next-token prediction pretraining. Our approach significantly improves transformer-based models, achieving impressive results in finance, e-commerce, and healthcare tasks.
\ifdefined\arxivpreprint
The code is publicly available on GitHub\footnote{\url{https://github.com/ivan-chai/pretpp}}.
\fi
\end{abstract}

\begin{figure}[t]
  \centering
\begin{subfigure}{\columnwidth}  
 \includegraphics[width=\columnwidth]{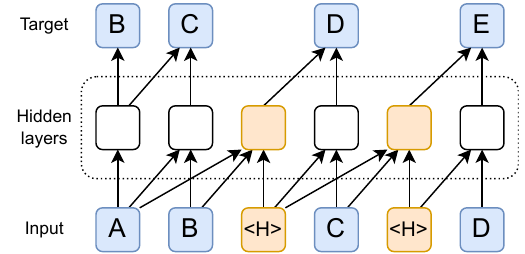}
 \caption{History Token serves as a bottleneck during pretraining.}
 \label{fig:general-pretrain}
\end{subfigure}
\par\medskip
\begin{subfigure}{\columnwidth}  
 \includegraphics[width=\columnwidth]{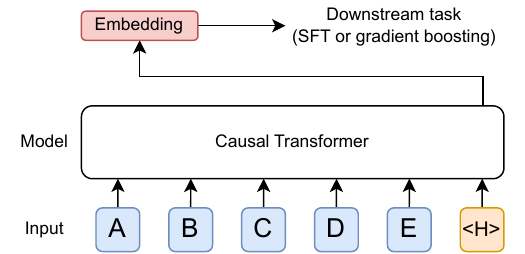}
 \caption{The embedding of History Token is used in downstream tasks.}
 \label{fig:general-downstream}
\end{subfigure}
\caption{History tokens accumulate prefix information during pretraining via next-token-prediction. The embedding of the History Token is later used in downstream tasks.}
\label{fig:general}
\end{figure}

\begin{figure*}[t]
  \centering
 \includegraphics{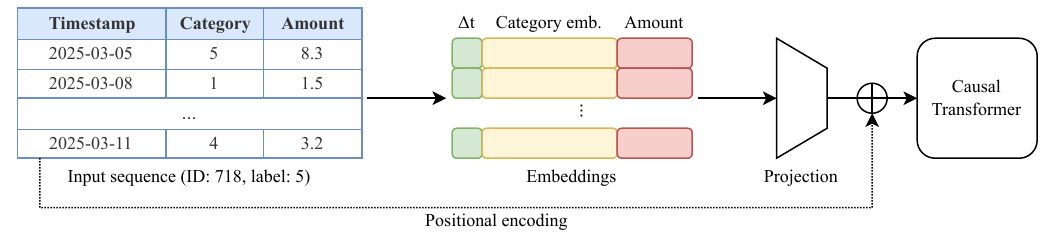}
\caption{Event sequences preprocessing.}
\label{fig:preprocessing}
\end{figure*}


\section{Introduction}

Many real-world problems involve predicting future events from historical observations. In continuation tasks, the goal is to forecast events that are similar to those previously observed~\cite{xueeasytpp}. However, many practical applications require anticipating events that do not explicitly appear in the training history. Examples include loan default, customer churn, and disease onset. These scenarios are typically addressed using classical machine learning models, such as logistic regression or gradient boosting, applied to handcrafted features or unsupervised model-based embeddings derived from historical data~\cite{osin2024ebes,recsys25challenge}.

In recent years, deep learning has shown significant success in modeling sequential structures, including event sequences, temporal point processes, and time series data. A prominent trend is the adoption of pretrained Transformer architectures due to their capacity to capture long-range dependencies and complex temporal patterns~\cite{padhi2021tabgpt,zuo2020thp}. Unlike recurrent neural networks, however, Transformers lack a canonical mechanism for extracting a fixed-size embedding from a sequence, as information is distributed across the activations of all tokens. This issue is commonly mitigated through auxiliary objectives during pretraining, such as contrastive learning~\cite{behnamghader2024llm2vec}, sentence order prediction~\cite{albert}, or next-sentence prediction~\cite{devlin2019bert}. However, each of these approaches introduces limitations. For instance, it is well documented that contrastive pretraining may overemphasize “easy features”, thereby compromising downstream quality~\cite{robinson2021easyfeatures}. Consequently, the problem of learning robust and informative sequence embeddings, including methods, based on the next-token prediction objective~\cite{yenduri2024gpt}, remains an open research question.

In this work, we propose a novel approach to pretraining Transformer-based embeddings without relying on auxiliary tasks. Our method draws inspiration from recurrent architectures and leverages sparse attention masks to guide the accumulation of historical information~\cite{bulatov2022recmemtransformer}. Specifically, we introduce \textit{History Tokens} that gather and summarize contextual information during training via a standard next-token prediction objective. We empirically evaluate the resulting embeddings across multiple domains, including finance, e-commerce, and healthcare, and show that they offer strong predictive performance, especially for future-oriented tasks, as opposed to global sequence classification.

The contributions of this paper are as follows:
\begin{enumerate}
\item We propose a novel HT-Transformer architecture that employs special history tokens to accumulate past information during pretraining using only the next-token prediction objective.
\item We develop advanced strategies for history tokens position selection and attention masking for improved downstream quality.
\item We demonstrate that the proposed method is particularly well-suited for predictive tasks focused on future events, as opposed to global sequence classification.
\item We establish new state-of-the-art results across a range of benchmarks in finance, e-commerce, and healthcare.
\end{enumerate}

\begin{figure*}[t]
  \centering
\begin{subfigure}[b]{0.24\textwidth}  
\centering
 \includegraphics[width=0.9\linewidth]{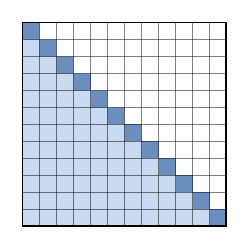}
 \caption{Causal.}
 \label{fig:mask-causal}
\end{subfigure}%
\begin{subfigure}[b]{0.24\textwidth}  
\centering
 \includegraphics[width=0.9\linewidth]{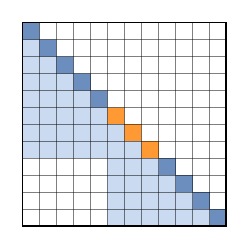}
 \caption{Recurrent Memory.}
 \label{fig:mask-rmtransformer}
\end{subfigure}
\begin{subfigure}[b]{0.24\textwidth}  
\centering
 \includegraphics[width=0.9\linewidth]{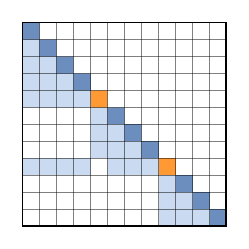}
 \caption{HT-Transformer (last).}
 \label{fig:mask-htlast}
\end{subfigure}
\begin{subfigure}[b]{0.24\textwidth}  
\centering
 \includegraphics[width=0.9\linewidth]{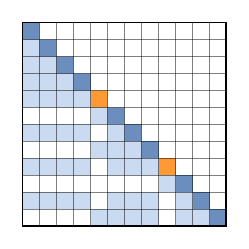}
 \caption{HT-Transformer (random).}
 \label{fig:mask-htrandom}
\end{subfigure}
\caption{Comparison of attention masks. Special tokens are orange-colored.}
\label{fig:mask}
\end{figure*}

\section{Preliminaries on Event Sequences}
This work focuses on modeling sequences of discrete events $S = \{s_i\}_{i=1}^N$, where each event $s_i$ is represented by a collection of fields, including a timestamp $t_i$, optional numerical attributes, and categorical variables. Each sequence typically corresponds to a single entity, such as a user or client, and the events are ordered chronologically by their timestamps: $t_1 \le t_2 \le \dots \le t_N$. An illustration of such sequences is provided in Figure~\ref{fig:preprocessing}.

\paragraph{Data preprocessing.} Before inputting data into a deep model, each event in the sequence must be transformed into an embedding in a latent space. In a typical preprocessing pipeline, each data field is encoded independently, and the resulting embeddings are concatenated to form a single event representation~\cite{gorishniy2021tabular}. Categorical features are transformed by assigning a trainable embedding vector to each possible value. Numerical features are incorporated directly into the event embedding without additional preprocessing.

When using Transformer models, we apply time-based positional encoding, following the approach proposed in prior work~\cite{yang2022attnhp}. Specifically, for each timestamp $t$, we compute a positional embedding $\mathrm{PE}_i(t)$ of dimension $d$ as:
\begin{equation}
    \mathrm{PE}_i(t) = \begin{cases}
    \sin\left(t / (m * (\frac{5M}{m})^\frac{i}{d})\right), & \text{if $i$ is even} \\
    \cos\left(t / (m * (\frac{5M}{m})^\frac{i - 1}{d})\right), & \text{otherwise}
    \end{cases}
\end{equation}
where $m$ and $M$ are constants determined from the distribution of timestamp values. For implementation details and parameter selection, we refer the reader to the original work.

\paragraph{Pretraining.} We consider sequence-level classification tasks, where each sequence $S$ is associated with a single target label $l$. Because labels are assigned at the sequence level rather than the event level, the number of labeled examples is often much smaller than the total number of events. This imbalance motivates the development of unsupervised pretraining algorithms that can leverage the abundance of unlabeled sequential data to improve downstream performance.

Unsupervised pretraining on event sequences is typically based on either generative or contrastive learning objectives.

In the \textit{generative} approach, the model is trained is trained to predict the next event $s_{i+1}$ given the historical context $s_1, \dots, s_i$, encouraging the model to capture temporal dependencies and sequence structure. A typical generative loss is formulated as a weighted sum of individual losses over each data field~\cite{shchur2019intensityfree,padhi2021tabgpt,mcdermott2023eventstreamgpt}. Timestamps can be predicted using standard regression losses such as Mean Absolute Error (MAE) or Mean Squared Error (MSE), or through more expressive temporal point process models based on intensity functions~\cite{rizoiu2017hawkes-tutorial,zuo2020thp}. In our work, we adopt the MAE loss for timestamp prediction:
\begin{equation}
\mathcal{L}_\mathrm{MAE}(\Delta\hat{t}, \Delta t) = |\Delta \hat{t} - \Delta t|,
\end{equation}
where $\Delta \hat{t}$ is the predicted inter-event time and $\Delta t$ is the ground truth. We apply the same MAE objective to other numerical fields and use the cross-entropy loss for categorical attributes.

An alternative to generative modeling is \textit{contrastive learning}~\cite{babaev2022coles}, which aims to learn sequence representations by maximizing agreement between different augmented views of the same sequence and pushing apart views from different sequences. Typically, each sequence is divided into multiple, possibly overlapping, subsequences $R_k \subset S$ for $k = 1, \dots, K$. Let $\mathrm{ID}(R)$ denote the index of the original sequence from which a chunk $R$ was derived. Then the contrastive loss~\cite{chopra2005contrastive} is defined as:
\begin{multline}
\mathcal{L}_\mathrm{cont}(R_i, R_j) = \\ \begin{cases}
\|f(R_i) - f(R_j)\|^2, & \text{if } \mathrm{ID}(R_i) = \mathrm{ID}(R_j) \\
\max\left(0, \epsilon - \|f(R_i) - f(R_j)\| \right)^2, & \text{otherwise}
\end{cases}
\end{multline}
where $f(R) \in \mathbb{R}^d$ is the embedding of a subsequence $R$ produced by the model. Following prior work~\cite{babaev2022coles}, we use $\epsilon = 0.5$ and $K = 5$ subsequences per sequence.

Both generative and contrastive paradigms have been successfully adapted to neural architectures such as recurrent neural networks (RNNs) and Transformers. However, while effective for certain tasks, these approaches exhibit notable limitations, especially when the goal is to anticipate future events rather than to summarize past behavior. Overcoming these limitations is a key motivation behind the approach proposed in this work.

\paragraph{Supervised fine-tuning.} An optional supervised fine-tuning (SFT) stage can be performed when labeled training data is available. During this phase, the output projection layer of the pretrained model is replaced with a new classification head designed to produce logits for the target task. The model is then trained using a standard cross-entropy loss over multiple epochs. To mitigate overfitting, additional regularization techniques such as Low-Rank Adaptation (LoRA) can be employed~\cite{hu2022lora}.

\section{Proposed Method}
The core idea of the proposed method is to introduce special \textit{history tokens} into Transformer models. These tokens are designed to accumulate information from preceding tokens in the sequence. A carefully constructed attention mask ensures that these tokens act as an information bottleneck, similar in function to the hidden states in recurrent neural networks (RNNs). In the following, we describe the training procedure and the application of history tokens for downstream classification tasks.

\subsection{Unsupervised Pretraining with History Tokens}

Transformer models for event sequences typically consist of three primary components: an event embedder, a backbone, and a prediction head. The proposed approach is compatible with any Transformer architecture that employs a causal attention mask, where each token attends only to preceding tokens.





History tokens are injected into the input of the backbone after event embeddings have been computed, as illustrated in Figure~\ref{fig:general-pretrain}. Each history token is assigned a timestamp equal to that of a preceding event, which is used for positional encoding.

To enable history tokens to serve as memory units, we modify the attention mask used by the backbone. Each history token is allowed to attend to all preceding event tokens (except other history tokens), thereby accumulating contextual information. In contrast, event tokens can attend only to history tokens and to event tokens occurring between the current position and the most recent history token. This attention pattern is illustrated in Figure~\ref{fig:mask-htlast}. When multiple history tokens are present, we introduce two attention strategies. In the \textit{Last} strategy, each event token is restricted to attend only to the most recent preceding history token. In the \textit{Random} strategy, illustrated in Figure~\ref{fig:mask-htrandom}, each event token selects one of the preceding history tokens at random during attention computation. As demonstrated in our experiments, the Random strategy yields consistently better performance across a range of tasks.

The proposed method allows considerable flexibility in selecting both the number and positions of history tokens.  For a given sequence of length $L$, the number of history tokens is computed as $\max(1, fL)$, where $f$ is a tunable hyperparameter referred to as the \textit{frequency}.

\begin{table*}[t]
\centering
\begin{tabular}{l|ccc|ccc|ccc}
\multirow{2}{*}{Dataset} & \multirow{2}{*}{\# Sequences} & \multirow{2}{*}{\# Events} & \multirow{2}{*}{\# Fields} & Mean & Mean & Time & \multicolumn{3}{c}{Downstream} \\
& & & & length & duration & unit & Target & \# Classes & Metric \\
\hline
Churn & 10217 & 1M & 6 & 99.3 & 80.5 & Day & Churn & 2 & ROC AUC \\
AgePred & 50000 & 44M & 3 & 875 & 718 & Day & Age group & 4 & Accuracy \\
Alfabattle & 1466527 & 343M & 15 & 234 & 275 & Day & Default & 2 & ROC AUC \\
MIMIC-III & 52103 & 23M & 3 & 407 & 108 & Day & Mortality & 2 & ROC AUC \\
Taobao & 9904 & 5M & 3 & 527 & 12.9 & Day & Activity & 2 & ROC AUC \\
\end{tabular}
\caption{Datasets statistics}
\label{tab:datasets}
\end{table*}

In our experiments, we compare two strategies for inserting history tokens into the sequence. The first inserts history tokens at uniformly sampled positions.The first places them at positions sampled uniformly across the sequence. However, this approach can lead to a discrepancy between training and inference, as history tokens are typically positioned near the end of the sequence during evaluation. To address this issue, we introduce the \textit{Bias-End} (BE) strategy, which places history tokens closer to the sequence's end. Specifically, it samples positions uniformly within the range $[\mu/2, L]$, where $\mu$ is the mean sequence length in the batch, and $L$ is the maximum sequence length. As our experiments show, the Bias-End strategy consistently leads to improved downstream performance.

At inference time, the history token is inserted only at the end of the sequence. In this setting, event tokens do not have access to any preceding history tokens, creating a mismatch with the pretraining setup, where history tokens may appear throughout the sequence. To mitigate this discrepancy, we apply history tokens in only a subset of pretraining batches with some \textit{application probability} $p$ (typically 50\%). This partial application encourages the model to remain robust across both configurations.

\subsection{Downstream classification}
During embedding extraction, a single history token is appended to the end of the input sequence, and the average of corresponding hidden activations from the Transformer backbone is used as the sequence-level embedding. This embedding can serve as an input feature for downstream models, such as gradient boosting classifiers. Alternatively, the entire Transformer model can be fine-tuned in a supervised setting by attaching a classification head to the output corresponding to the history token. In our experiments, we evaluate both approaches: using frozen embeddings as input to external models and fine-tuning the Transformer end-to-end for classification.

\section{Related Work}
Transformer models have a long and successful history of application to sequence classification tasks, particularly in natural language processing (NLP)~\cite{vaswani2017attention}. One of the main challenges in this setting is the limited availability of labeled data, which has driven the development of effective unsupervised pretraining strategies~\cite{muennighoff2023mteb}. A notable early approach is BERT~\cite{devlin2019bert}, which introduced a masked language modeling (MLM) objective alongside a next sentence prediction (NSP) task to enable powerful sequence representations. These pretrained models proved highly effective for downstream classification tasks such as natural language understanding (NLU)~\cite{glue}.

A central issue in applying Transformers to NLU is how to extract a compact, semantically meaningful representation of an entire sequence. In BERT, this was addressed by introducing a special classification token trained via the NSP objective. However, subsequent work such as RoBERTa~\cite{liu2019roberta} challenged the necessity of the NSP task, demonstrating that it could be omitted without degrading performance.

Beyond BERT-style objectives, other works have explored contrastive pretraining techniques, such as LLM2Vec~\cite{behnamghader2024llm2vec}, which aim to bring semantically similar sequences closer in embedding space. While contrastive learning can yield strong performance when carefully implemented, it suffers from notable limitations. In particular, models can rely on “easy” features, such as surface-level similarities, to distinguish positive pairs, bypassing the need for deeper semantic understanding. Furthermore, contrastive pretraining tends to bias models toward capturing global sequence properties at the expense of local or up-to-date information. This bias poses a particular challenge in event sequence modeling, where the most recent context is often essential for accurate prediction, in contrast to many natural language processing tasks that emphasize global semantics.

\begin{table*}[t]
\centering
\begin{tabular}{p{4cm}|ccccc}
    Method & Churn & AgePred & Alfabattle & MIMIC-III & Taobao \\ 
    \hline
    NTP RNN & 81.56 $\pm$ 0.59 & 60.05 $\pm$ 0.29 & 79.83 $\pm$ 0.05 & 90.68 $\pm$ 0.07 & 83.28 $\pm$ 1.42 \\ 
    NTP Transformer & 80.92 $\pm$ 0.66 & 56.16 $\pm$ 0.51 & 78.63 $\pm$ 0.12 & 91.28 $\pm$ 0.10 & 83.39 $\pm$ 1.43 \\ 
    CoLES RNN & 82.82 $\pm$ 0.28 & \bf 62.42 $\pm$ 0.33 & 79.30 $\pm$ 0.08 & 87.44 $\pm$ 0.20 & \bf 85.56 $\pm$ 1.14 \\ 
    CoLES Transformer & 78.92 $\pm$ 0.49 & 59.92 $\pm$ 0.30 & 78.40 $\pm$ 0.00 & 87.06 $\pm$ 0.38 & 82.03 $\pm$ 0.98 \\ 
    \hline
    HT-Transformer & \bf 83.34 $\pm$ 0.42 & 60.10 $\pm$ 0.39 & \bf 80.42 $\pm$ 0.12 & \bf 92.00 $\pm$ 0.09 & 84.65 $\pm$ 1.07 \\
\end{tabular}
\caption{Pretrained models classification results.}
\label{tab:results-es-class-pretrain}
\end{table*}

\begin{table*}[t]
\centering
\begin{tabular}{p{4cm}|ccccc}
    Method & Churn & AgePred & Alfabattle & MIMIC-III & Taobao \\ 
    \hline
    Supervised RNN & 79.10 $\pm$ 0.80 & 61.18 $\pm$ 0.49 & 76.47 $\pm$ 1.13 & 91.46 $\pm$ 0.10 & 84.91 $\pm$ 1.17 \\ 
    Supervised Transformer & 80.92 $\pm$ 0.66 & 54.88 $\pm$ 2.37 & 74.90 $\pm$ 0.08 & 77.48 $\pm$ 1.22 & 79.71 $\pm$ 1.68 \\ 
    \hline
    NTP RNN + SFT & 82.80 $\pm$ 0.40 & 61.07 $\pm$ 0.86 & 80.27 $\pm$ 0.12 & 91.82 $\pm$ 0.07 & 85.03 $\pm$ 2.64 \\ 
    NTP Transformer + SFT & 82.52 $\pm$ 0.19 & 64.09 $\pm$ 0.31 & \bf 81.70 $\pm$ 0.17 & 92.91 $\pm$ 0.15 & 86.12 $\pm$ 1.02 \\
    CoLES RNN + SFT & 82.04 $\pm$ 0.63 & 63.26 $\pm$ 0.58 & 79.00 $\pm$ 0.14 & 89.88 $\pm$ 0.51 & 86.36 $\pm$ 0.46 \\ 
    CoLES Transformer + SFT & 80.68 $\pm$ 0.53 & 60.91 $\pm$ 0.51 & 80.82 $\pm$ 0.12 & 84.43 $\pm$ 4.25 & 81.81 $\pm$ 1.93 \\
    \hline
    HT-Transformer + SFT & \bf 83.76 $\pm$ 0.50 & \bf 64.26 $\pm$ 0.30 & 81.63 $\pm$ 0.05 & \bf 92.97 $\pm$ 0.07 & \bf 87.29 $\pm$ 0.52 \\ 
\end{tabular}
\caption{Fine-tuned models classification results.}
\label{tab:results-es-class-sft}
\end{table*}

Alternative methods for sequence embedding extraction include averaging token activations across certain layers or using the final token’s activation~\cite{stankevivcius2024extractingembeddings}. However, these approaches generally underperform compared to specialized embedding pretraining techniques, particularly in tasks requiring nuanced or fine-grained representations.

In contrast, recurrent neural networks (RNNs) provide a natural mechanism for summarizing sequences, as the hidden state at the final timestep inherently encodes the information required for future prediction. This property has motivated the incorporation of recurrent principles into Transformer architectures, particularly for modeling long sequences~\cite{bulatov2022recmemtransformer}. Related ideas also appear in architectures like Longformer~\cite{beltagy2020longformer}, where global tokens are used to aggregate and propagate information across extended contexts. More recently, recurrent-style Transformers have been combined with contrastive learning objectives to achieve strong performance on natural language understanding (NLU) tasks, while preserving the generative capabilities of causal models~\cite{zhang2025gem}.

Our work extends the Recurrent Transformer paradigm~\cite{bulatov2022recmemtransformer} by introducing a novel mechanism for representation learning from event sequences. We propose the use of history tokens, which are designed to accumulate and summarize historical context during next-token prediction (NTP) pretraining. Experimental results demonstrate that history tokens are particularly effective for forecasting future events, a task that is central to many real-world applications but rarely addressed in standard NLP settings.


\section{Experiments}
We conduct experiments on datasets spanning multiple domains. The Churn\footnote{\url{https://boosters.pro/championship/rosbank1/}}, AgePred\footnote{\url{https://ods.ai/competitions/sberbank-sirius-lesson}}, and Alfabattle\footnote{\url{https://boosters.pro/championship/alfabattle2/overview}} datasets represent a range of downstream tasks in the financial technology domain. MIMIC-III~\cite{johnson2016mimic} is a widely used collection of medical records, and the Taobao dataset\footnote{\url{https://tianchi.aliyun.com/dataset/46}} represents user interactions in e-commerce. Summary statistics for these datasets are provided in Table~\ref{tab:datasets}.

We evaluate three primary baseline approaches: supervised learning, next-token prediction (NTP)~\cite{radford2018decoderonly}, and contrastive learning using CoLES~\cite{babaev2022coles}. Each method is applied to two backbone architectures. For RNN-based models, we use a GRU backbone~\cite{cho2014gru}, while Transformer-based models employ a decoder-only architecture~\cite{radford2018decoderonly}.

All models are trained using the Adam optimizer~\cite{kingma2014adam} with a fixed learning rate of 0.001. The maximum number of training epochs varies by dataset and ranges from 60 to 120. Early stopping is applied based on validation performance to prevent overfitting. Supervised fine-tuning is performed for 20 epochs.

Experiments were conducted on NVIDIA A100 GPUs. For all datasets except Alfabattle, training was performed on a single GPU. Due to the larger size of the Alfabattle dataset, some experiments were accelerated using 2 GPUs to reduce training time.

Hyperparameters, including the loss weights for the NTP objective and model size, are optimized using a Bayesian optimizer~\cite{snoek2012bayesian} applied to the NTP RNN baseline. The resulting configurations are reused across all other RNN settings. For Transformer models, we separately tune the number of layers and the hidden dimension using the NTP configuration and apply these settings consistently across all Transformer-based variants.



For each method, we report the mean and standard deviation of evaluation metrics across five different random seeds. An exception is made for the Alfabattle dataset, where three seeds were used due to computational constraints.


To assess the quality of extracted embeddings, we train a gradient boosting classifier for each downstream task using the LightGBM library~\cite{ke2017lightgbm}. The classifier is trained on frozen embeddings and uses the same hyperparameters as in the CoLES baseline~\cite{babaev2022coles}.

\begin{figure*}[t]
\begin{minipage}[b]{0.65\textwidth}
    \centering
    \captionsetup{type=figure}
    \includegraphics[width=\linewidth]{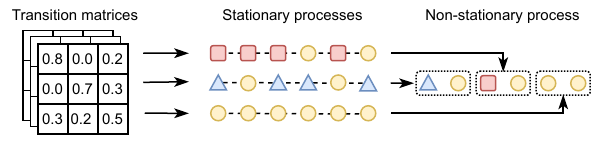}
    \caption{Markovian generative process for the toy dataset.}
    \label{fig:toy-generation}
  \end{minipage}
  \hfill
  \begin{minipage}[b]{0.34\textwidth}
    \captionsetup{type=table}
    \centering
    \begin{tabular}{l|cc}
    Method & \thead{Local \\ (Last part)} & \thead{Global \\ (Num. parts)} \\
    \hline
    Supervised & 0.71 & 1.00 \\
    \hline 
    NTP Last & 0.53 & 0.73 \\
    NTP Avg & 0.40 & 0.88\\
    CoLES & 0.33 & \bf 0.94 \\
    \hline
    NTP HT & \bf 0.55 & 0.85 \\
    \end{tabular}
    \caption{Toy dataset classification accuracy.}
    \label{tab:toy}
  \end{minipage}
\end{figure*}

\subsection{Classification of Event Sequences}
Classification results are presented in Table~\ref{tab:results-es-class-pretrain} for models after unsupervised pretraining and in Table~\ref{tab:results-es-class-sft} following supervised fine-tuning (SFT).

Among baselines in the unsupervised setting, the standard NTP Transformer significantly outperforms its RNN counterpart only on the MIMIC-III dataset, while performing worse on Churn, AgePred, and Alfabattle. This highlights the limitations of traditional Transformer architectures in learning compact and informative sequence representations for downstream tasks.

The proposed HT-Transformer effectively addresses these limitations. It consistently outperforms the NTP Transformer and achieves the best results across all evaluations, with the exception of supervised fine-tuning on the AlfaBattle dataset. In the pretraining setting, HT-Transformer delivers superior performance on three datasets; however, on AgePred and Taobao, its embeddings underperform compared to those produced by the RNN-based CoLES method. Nevertheless, after supervised fine-tuning, HT-Transformer surpasses all baselines on these datasets.

The AgePred task is different from the others because it requires predicting a global property, specifically the client’s age group, using historical event data. As discussed in the following section, history tokens are designed to capture recent and predictive information, which is beneficial for forecasting future events but less effective for tasks that require encoding global sequence properties. As a result, embeddings extracted from HT-Transformer are less suitable for such tasks without additional supervision.

In summary, the proposed method consistently outperforms other Transformer-based approaches and achieves state-of-the-art results across all evaluated datasets after supervised fine-tuning. These findings demonstrate the effectiveness of HT-Transformer for future-oriented prediction tasks and highlight its adaptability to global classification problems when supervised fine-tuning is applied.





\subsection{Global Classification and Future-Oriented Tasks}
While the concept of history tokens is broadly applicable, we observe certain limitations when they are used in combination with next token prediction during pretraining. The next token prediction objective encourages the model to focus on extracting recent information that is directly relevant for forecasting upcoming events. In contrast, downstream tasks involving classification based on global or persistent properties, such as long-term user characteristics, may benefit more from contrastive pretraining or from simpler aggregation strategies, such as averaging Transformer outputs across the sequence.

To investigate this effect, we conduct experiments on a synthetic dataset specifically designed to evaluate the suitability of different representation learning methods for local versus global tasks. In this dataset, we sample ten distinct transition matrices, each defining a Markov process by specifying the probability of transitioning from one label to another. We then construct nonstationary sequences by concatenating multiple segments, each generated using a different transition matrix, as illustrated in Figure~\ref{fig:toy-generation}.

We introduce two classification tasks for our synthetic dataset. The global classification task requires predicting the total number of transition matrices used in a sequence, which ranges from 1 to 5. This task demands that the model capture information across the entire sequence. In contrast, the local classification task involves identifying the index of the transition matrix used in the final segment, which depends only on the most recent data.

Results of classification experiments using Transformer-based models are presented in Table~\ref{tab:toy}. These results support the conclusion that History Tokens are particularly well suited for tasks that rely on recent context, such as future event prediction. On the other hand, contrastive pretraining and embedding averaging are more effective for global classification tasks that require holistic sequence understanding.



\begin{table}[t]
\centering
\begin{tabular}{l|ccc|c}
    Method & Churn & MIMIC & Taobao & AVG \\
    \hline
    Uniform pl. & 83.23 & 91.92 & 83.72 & 86.29 \\
    + SFT & 82.88 & \bf 93.05 & 84.63 & 86.85 \\
    Last sel. & 82.92 & 91.90 & 83.78 & 86.20 \\
    + SFT & 82.69 & 93.01 & 85.75 & 87.15 \\
    \hline
    HT-Transformer & 83.34 & 92.00 & 84.65 & 86.66 \\
    + SFT & \bf 83.76 & 92.97 & \bf 87.29 & \bf 88.01  \\
\end{tabular}
\caption{Comparison of history token placement and selection strategies.}
\label{tab:ablation-strategy}
\end{table}


\subsection{Ablation Studies}
\paragraph{Training Strategies.} In the introduction of HT-Transformer, we outlined alternative strategies for history token placement and selection. Table~\ref{tab:ablation-strategy} compares these alternatives with the default HT-Transformer configuration, which employs the Bias-End placement strategy and Last selection of history tokens. As shown, the default configuration yields superior downstream performance on the Churn and MIMIC datasets. However, on MIMIC-III, it results in slightly lower performance compared to one of the alternatives. Overall, the results indicate that both the placement and selection strategies have a significant impact and contribute meaningfully to the final model quality.

\paragraph{Hyper-parameters.}
HT-Transformer introduces two key hyper-parameters: the insertion frequency $f$ of history tokens and the application probability $p$. The insertion frequency determines the number of history tokens relative to the input length, while the application probability specifies the proportion of training batches in which history tokens are applied.

\begin{figure}[t]
\centering
\includegraphics[width=\linewidth]{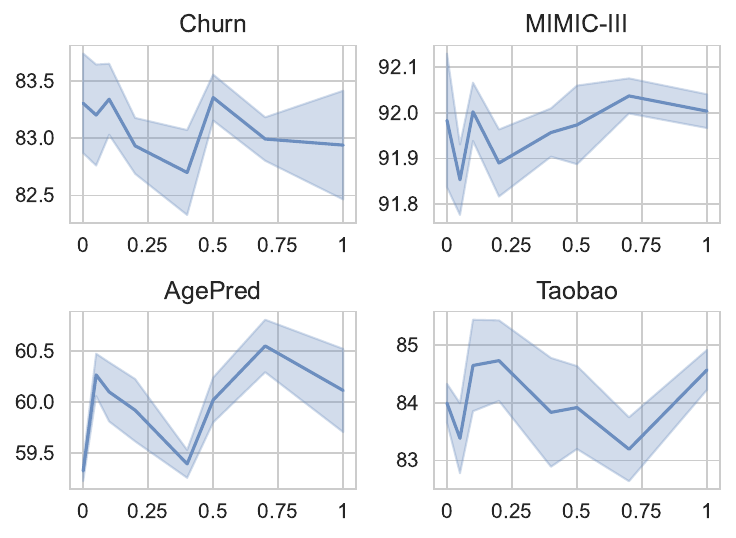}
\caption{The dependency of the pretraining quality on the history token insertion frequency $f$. Zero frequency corresponds to a single token per sequence.}
\label{fig:ablation-f}
\end{figure}

Figure~\ref{fig:ablation-f} shows that the model's performance does not strongly depend on the exact value of $f$. In most cases, even a single history token achieves comparable performance to configurations with more frequent insertion. The only notable exception is the AgePred dataset, where increasing the number of history tokens leads to improved performance.

\begin{figure}[t]
\centering
\includegraphics[width=\linewidth]{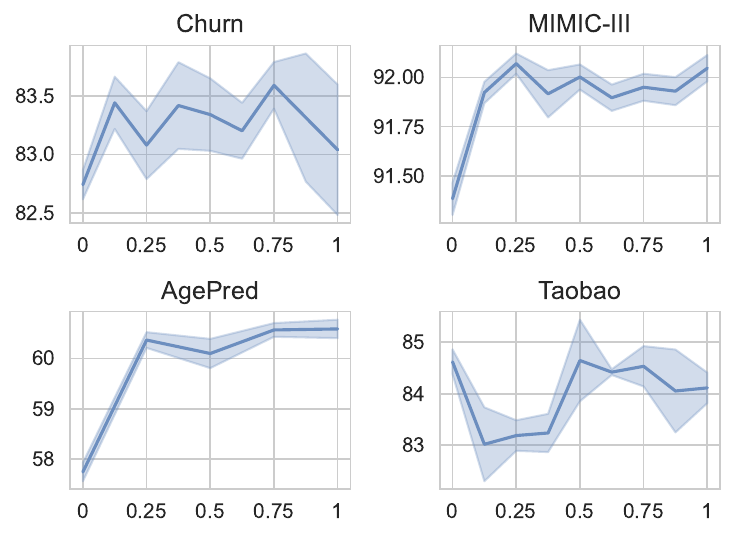}
\caption{The dependency of the pretraining quality on the history token application probability $p$.}
\label{fig:ablation-p}
\end{figure}

Figure~\ref{fig:ablation-p} illustrates the effect of varying the application probability $p$. The results indicate that setting $p$ too low significantly degrades performance. On the Churn dataset, using the maximum value $p = 1$ also results in a modest performance drop. Interestingly, training with $p = 0$ still outperforms a standard NTP Transformer. Our analysis revealed that using the embedding of a randomly initialized $[\mathrm{CLS}]$ token at the end of the sequence performs better than using the final token's output representation.

Based on these observations, we recommend setting the history token frequency to 10\% of the input length and the application probability $p$ to 50\%, as used in our default configuration. This setting provides the most stable and consistent performance across all evaluated datasets.



\section{Limitations and Future Work}
In this paper, we demonstrated the effectiveness of using history tokens for event sequence classification. We introduced a new Transformer-based architecture, evaluated multiple design choices, and identified configurations that lead to strong downstream performance across a range of domains. However, several aspects of the method remain open for further exploration.

First, as shown in our ablation studies, the placement of history tokens has a significant impact on downstream performance. We compared two strategies: uniform placement and the Bias-End approach. A more detailed analysis of token positioning and sampling policies could be pursued in future work.

Second, our current implementation relies on the standard PyTorch multi-head attention module. This component may not be optimal for working with the custom attention masks required by the HT-Transformer. Future technical improvements could focus on optimizing the attention mechanism, particularly by exploiting the sparsity of the mask. Since only a small subset of tokens participates in the full self-attention computation, the total computational cost can be reduced. As a result, the HT-Transformer has the potential to offer faster training compared to conventional causal Transformers.

Overall, the proposed architecture offers a promising direction for modeling event sequences with high efficiency and accuracy. We believe that future work can continue to improve both the predictive performance and computational scalability of the method.

\section{Conclusion}
This paper introduced HT-Transformer, a novel architecture designed to enhance Transformer-based models for event sequence classification by explicitly accumulating historical information through learnable history tokens. We identified the inherent limitations of standard Transformers in tasks requiring future event prediction, specifically the lack of a unified representation that captures sequential context effectively. To address this limitation, we proposed a simple yet effective mechanism where history tokens act as information bottlenecks during next-token prediction pretraining, analogous to hidden states in recurrent neural networks.

Our method eliminates the need for auxiliary objectives such as contrastive learning, instead leveraging sparse attention patterns to ensure efficient information aggregation. Extensive empirical evaluations across real-world datasets from finance, healthcare, and e-commerce demonstrated that HT-Transformer consistently outperforms conventional Transformer baselines. After supervised fine-tuning, the model achieved state-of-the-art results in 4 out of 5 benchmarks, with improvements of up to 0.96\% in ROC AUC on the Churn dataset.

Overall, HT-Transformer represents a significant step forward in bridging the performance gap between recurrent and Transformer-based models for future-oriented sequence modeling. 




\bibliography{main}

\end{document}